\documentclass{bmvc2k}

\usepackage{amsmath}
\usepackage{bbm}
\usepackage{fdsymbol}
\usepackage{color, colortbl}

\usepackage{pifont}

\title{Learning Temporal Sentence Grounding From Narrated EgoVideos}

\addauthor{Kevin Flanagan}{Kevin.Flanagan@bristol.ac.uk}{1}
\addauthor{Dima Damen}{Dima.Damen@bristol.ac.uk}{1}
\addauthor{Michael Wray}{Michael.Wray@bristol.ac.uk}{1}

\addinstitution{
 School of Computer Science\\
 University of Bristol\\
 Bristol, UK
}

\runninghead{Flanagan et al}{Learning Temporal .. From Narrated EgoVideos}

\def\eg{\emph{e.g}\bmvaOneDot}
\def\Eg{\emph{E.g}\bmvaOneDot}
\def\ie{\emph{i.e}\bmvaOneDot}

\def\etal{\emph{et al}\bmvaOneDot}

\newcommand{\methodName}{Cli$\mathcal{M}$er}
\newcommand{\taskName}{Temporal Sentence Grounding}
\newcommand{\taskNameMini}{TSG}

\newcommand{\myparagraph}[1]{\vspace{2pt}\noindent\textbf{#1}}
\newcommand{\refSup}[1]{\textcolor{red}{#1}}

\begin{document}

\maketitle

\begin{abstract}
The onset of long-form egocentric datasets such as Ego4D and EPIC-Kitchens presents a new challenge for the task of Temporal Sentence Grounding (TSG). Compared to traditional benchmarks on which this task is evaluated, these datasets offer finer-grained sentences to ground in notably longer videos. 
In this paper, we develop an approach for learning to ground sentences in these datasets using only narrations and their corresponding rough narration timestamps.
We propose to artificially merge clips to train for temporal grounding in a contrastive manner using text-conditioning attention.
This Clip Merging (\methodName{}) approach is shown to be effective when compared with a high performing TSG method---e.g. mean R@1 improves from 3.9 to 5.7 on Ego4D and from 10.7 to 13.0 on EPIC-Kitchens.
Code and data splits available from: \url{https://github.com/keflanagan/CliMer}
\end{abstract}

\begin{figure}[ht]
    \centering
    \includegraphics[width=\textwidth]{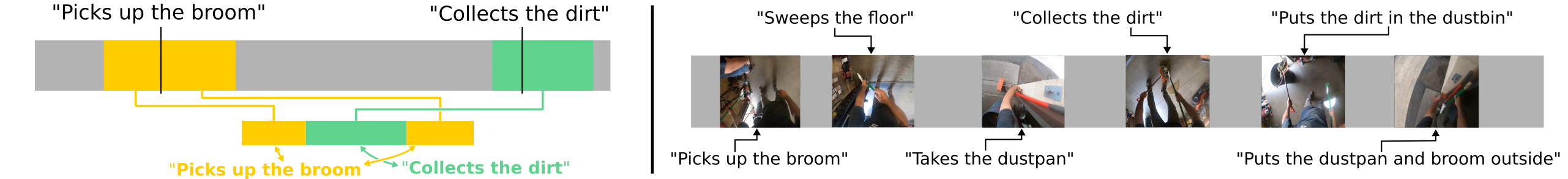}
    \caption{Left: We generate training examples from videos with rough timestamps of narrations by artificially merging clips to provide a contrastive signal. Right: At test time, \methodName{} can perform temporal grounding in long, dense videos.}
    \vspace*{-12pt}
    \label{fig:intro_fig}
\end{figure}

\section{Introduction}
\label{sec:intro}

Suppose you have recorded a video of your birthday party using a wearable device, and you wish to find the exact moment you blew out the candles, or ripped open a present with reckless abandon.
Searching through video data to retrieve the temporal extents of meaningful moments can be an arduous task.
A method that can take in a sentence describing the intended moment and return that moment, as a short temporal segment, is known as \taskName~\cite{anne2017localizing,gao2017tall}. This problem is commonly explored within datasets with videos of a couple of minutes duration, typically grounding clips 10s of seconds long~\cite{krishna2017anet,gao2017tall,anne2017localizing,regneri2013tacos}.

In this work, we explore \taskName{} (TSG) for long-form egocentric video datasets where searches may correspond to segments of only a few seconds in videos that are up to an hour in length  (See Table~\ref{tab:datasets}).
Recent works in \taskNameMini{} have explored long-form videos~\cite{barrios2023multimodal,hou2021cone}, yet still require full supervision in the form of start/end times of clips.
In contrast, rough timestamps from video narrations have previously been used for learning other tasks such as recognition~\cite{moltisanti2019action}, detection~\cite{li2021temporal,ma2020sf,rahaman2022generalized}, and spatio-temporal localisation~\cite{mettes2016spot}. 
In~\cite{ma2020sf}, Ma~\etal found that narration timestamps roughly located at or near the relevant action require $6\times$ less annotation effort than labelling with start/end times. The improved efficiency is crucial for event-rich datasets with long videos.
This proposes the unique problem that we investigate within this paper: how do we train a successful model for \taskNameMini{} on long videos with only narrations and their rough timestamps as supervision.

Our proposed method, \methodName, artificially generates training examples from long-form narrated videos by merging clips together to form a merged segment, Figure~\ref{fig:intro_fig} (left).
The features from the merged segment can be conditioned with the narrations, thus providing hard boundaries for \methodName{} to learn from in a contrastive learning set-up.
Because of this training setup, at inference time our method is able to perform \taskNameMini{} over long-form videos with a high density of annotated sentences, Figure~\ref{fig:intro_fig} (right).

To summarise, (i) we propose \methodName{} for \taskName{}, only using narrations and their rough timestamps as supervision.
\methodName{} merges clips as a supervisory signal to use for contrastive learning.
(ii) We explore for the first time using Ego4D and EPIC-Kitchens for \taskNameMini{} providing train-test splits for this task.
(iii) We show that \methodName{} outperforms the baseline VSLNet~\cite{zhang2021natural} on these two datasets, ablating our design decisions.

%

\vspace*{-6pt}
\section{Related Work}
\label{sec:related_work}

\begin{table}[t]
\centering
\resizebox{1\linewidth}{!}{
\begin{tabular}{lrrrrrr}
\hline
 & \begin{tabular}{@{}c@{}}Total Vid \\ Duration\end{tabular} 
 & \begin{tabular}{@{}c@{}}Avg. Vid \\ Duration\end{tabular}                  & \begin{tabular}{@{}c@{}}Avg. Mom. \\ Duration\end{tabular}         &  \begin{tabular}{@{}c@{}}Annotations \\ / Video\end{tabular}  & \begin{tabular}{@{}c@{}}Total \\ Annotations\end{tabular} & \begin{tabular}{@{}c@{}}Avg. \\ Coverage $\downarrow$\end{tabular}\\ \hline
ANet-Captions~\cite{krishna2017anet}   & 487.6h   & 2.0min   & 37.1s &  4.9   & 72k &  30.90\%  \\ 
Charades-STA~\cite{gao2017tall}        & 57.1h    & 0.5min   & 8.1s  &  2.3   & 16k &   27.00\% \\ 
DiDeMo~\cite{anne2017localizing}       & 88.7h    & 0.5min   & 6.5s  &  3.9   & 41k &   21.70\% \\
TACoS~\cite{regneri2013tacos}          & 10.1h    & 4.8min   & 27.9s &  143.6 & 18k &   9.70\%  \\ \hline
Ego4D~\cite{grauman2022ego4d}          & 234.9h   & 17.3min  & 2.0s  & 214.1  & 223k&   \textbf{0.19\%}  \\
EPIC-Kitchens~\cite{damen2022rescaling}& 73.4h    & 8.9min   & 3.1s  &  134.1 & 67k &  \underline{0.58\%}   \\ \hline
\end{tabular}}
\caption{Temporal Sentence Grounding Datasets---comparative statistics. We compare previously used datasets (top) to the egocentric ones we use (bottom). Avg. Coverage showcases the ratio of the ground-truth moment duration to the video duration. Ego4D  and EPIC-Kitchens have the shortest avg. moment duration with significantly lower coverage ($< 1\%$).}
\label{tab:datasets}
\end{table}







\myparagraph{\taskNameMini{} Datasets}
Table \ref{tab:datasets} shows a comparison between datasets previously used for \taskName{} and the two EgoVideo datasets we explore in this work, Ego4D~\cite{grauman2022ego4d} and EPIC-Kitchens~\cite{damen2022rescaling}.
Avg. Coverage (Avg. of Moment Duration/Video Duration) in particular displays the distinction between these datasets. Previous datasets aim to retrieve a moment $\ge 10\%$ of the video length on average. 
Ego4D and EPIC-Kitchens have a significantly smaller coverage ($0.19\%$ and $0.58\%$ respectively), due to their fine-grained sentences with much shorter average moment duration. 
Additionally, biases within previous TSG datasets (\cite{gao2017tall,krishna2017anet}) have been explored in~\cite{otani2020uncovering, yuan2021closer}. Specifically,~\cite{yuan2021closer} finds that the evaluated methods ``fail to utilize the video temporal relation or vision language interaction'' and was shown to be even more catastrophic for weakly supervised approaches.

Our motivation to explore datasets with significantly smaller coverage is related to the seminal dataset of movie descriptions, MAD~\cite{soldan2022mad}. In MAD, movies up to 3 hours long were explored for temporal grounding.
Movies offer different challenges including plot understanding and they typically contain scene boundaries. In this work, we focus on EgoVideos that are unedited with finer-grained narrations.

\myparagraph{Fully Supervised Grounding}
Fully supervised approaches make use of exact start and end times for each sentence during training.
Approaches are divided between proposal-based and proposal-free.
Proposal-based methods~\cite{anne2017localizing,cao2021pursuit,gao2017tall,ge2019mac,jiang2019cross,liu2018cross,liu2021context,liu2022exploring,xu2019multilevel,chen2018temporally,qu2020fine,wang2020temporally,zhang20192dtan} generate a set of candidate segments within the video and rank these for a given sentence.
Proposal-free methods~\cite{chen2020learning,chen2020hierarchical,chen2020rethinking,chen2021end,hao2022query,liu2022memory,lu2019debug,nan2021interventional,rodriguez2020proposal,yuan2019find,zhang2021natural,zhao2020bottom} directly predict the start and end times.
A small number of works have tackled the task of fully supervised long video grounding~\cite{hou2021cone, barrios2023multimodal}. These have taken the approach of using a secondary module to first pick out segments from the full video before applying models better suited to short video grounding within these segments. 

\myparagraph{Weakly Supervised Grounding} Weakly supervised approaches~\cite{chen2020look,chen2022intercon,cui2022viga,lin2020weakly,mithun2019weakly,wang2021weakly,wenfei2021lcnet,xu2022point,zheng2022cpl,zheng2022weakly} do not have access to start and end times for each sentence during training.
Instead, videos are paired with sentences which can be grounded somewhere within them.
Recent approaches include CPL~\cite{zheng2022cpl}, which generates content-dependent proposals and mines hard negative samples from within the same video, and LCNet~\cite{wenfei2021lcnet}, which extracts a hierarchical feature representation for video and text and models the local correspondences between these.
Similar to our work,~\cite{chen2022intercon} also creates merged videos, yet their merging approach assumes a dataset with high coverage by sampling a random clip from the video to act as the ground truth for a sentence.
In our method, we merge clips based on the rough timestamp supervision and use hard negative mining to ensure a trainable signal for Ego4D and EPIC-Kitchens, which have much lower coverage. 
Other recent approaches~\cite{cui2022viga, xu2022point} have made use of ``glance''/``point'' annotations, which are similar to the rough timestamp we use, but both works make the assumption that the ``glance''/``point'' is within the ground truth segment which can be unrealistic given potential annotation noise, either human or automatic~\cite{miech2019howto100m}.
\vspace*{-6pt}
\section{Method}
\label{sec:method}

We first introduce the task of Temporal Sentence Grounding (TSG) from Narration Timestamp Supervision in Sec.~\ref{subsec:wstg}. Next, we provide an overview of the method in Sec.~\ref{subsec:method_overview}.
Finally, we give details of sampling sentences and creating the merged segments in Sec.~\ref{subsec:clip_combine} followed by
text conditioning in Sec.~\ref{subsec:text_condition}, as well as training losses and inference in Sec.~\ref{sec:loss}.

\subsection{Task Formulation: TSG from Narration Timestamps}
\label{subsec:wstg}

Temporal Sentence Grounding is the problem of finding moments of a video that ground a sentence.
Formally, given a sentence $c_i$ and a video $x$\footnote{For simplicity we drop the video index as \taskName{} assumes oracle knowledge of the video that the sentence corresponds to.}, we wish to find the start and end times---given by $t^s_i$ and $t^e_i$ respectively---within $x$ which ground the sentence $c_i$.

In the fully supervised setting, during both training and testing, methods use the tuple $(x, c_i, t^s_i, t^e_i)$ to first train and later evaluate the model by comparing predicted times ($\hat{t}^s_i$ and $\hat{t}^e_i$) with the ground truth times.
In this work, we explore a weaker form of supervision in which only rough timestamps are available for each sentence during training.
These represent a narration's single timestamp $t_i$ that roughly corresponds to the sentence as labelled by a narrator.
Thus, in this setting, models are trained using $(x, c_i, t_i)$ but are still evaluated on their ability to ground the sentence to its annotated start/end times.
Naturally, this leads to a more difficult training regime, yet has the benefit of requiring much less annotation time~\cite{ma2020sf}.


\subsection{\methodName{} Overview}
\label{subsec:method_overview}
\begin{figure*}[t]
    \centering
    \includegraphics[width = \textwidth]{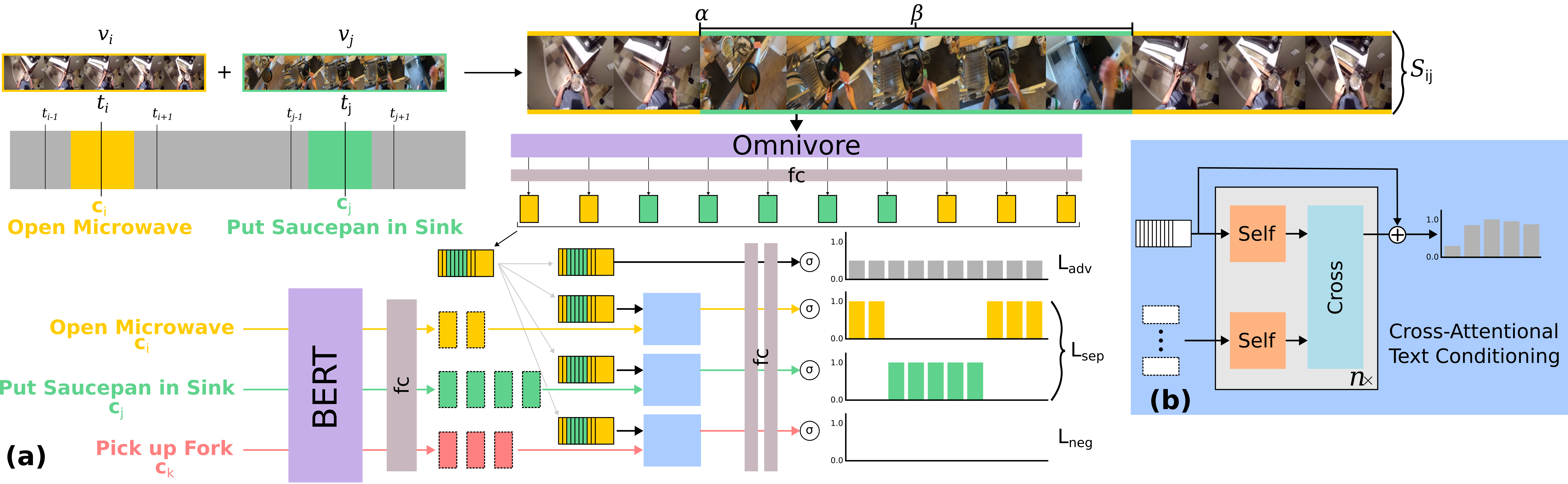}
    \caption{Overview of \methodName. (a) We generate a merged segment ($S_{ij}$) from two clips ($v_i$/$v_j$) of one long video to contrast against sentences $c_i$/$c_j$/$c_k$. The video features are conditioned using extracted text features and trained with 3 losses $L_{adv}, L_{sep}, L_{neg}$. (b) Cross-Attentional Text Conditioning uses $n$ blocks
 of self- and cross-attention to condition the text.}
 \vspace*{-12pt}
    \label{fig:model_full}
\end{figure*}

An overview of the approach can be seen in Figure~\ref{fig:model_full}, where from an untrimmed video two sentences $c_i$ (yellow) and $c_j$ (green) are sampled. We additionally sample a third sentence from the same video, $c_k$ (red), as a negative sentence in order to train the model to distinguish video segments that do not contain the corresponding grounding.
As discussed in the previous section, during training we only have access to single rough timestamps from the narrations as supervision, \eg $t_i$/$t_j$, for sentence $c_i$/$c_j$.
We first generate rough clips, $v_i$ and $v_j$ around $t_i$ and $t_j$.

Next, we create a new input $S_{ij}$ for training, which we term the merged segment, by merging the clips $v_i$ and $v_j$.
This is beneficial for two reasons: firstly, the artificial boundaries between clips act as a supervision signal; and secondly, contrasting multiple sentences ensures that the method learns to discriminate and ground individual sentences.
As we artificially perform the merging, we use the merging signal as supervision for learning to ground.

We condition each sentence against $S_{ij}$, using cross-attentional text conditioning (Figure~\ref{fig:model_full}~(b)) and train the model with three losses which contrast the merged segment with the three sentences and regularise the predictions.
We detail our method next.

\subsection{Sampling Sentences and Merging Segments}
\label{subsec:clip_combine}
We now explain how to sample the sentences, from a given video, and create a merged segment on which to train.
We avoid neighbouring clips, by ensuring \eg $|i-j| \geq 3$.
Additionally, we check that all three sentences $c_i$, $c_j$, and $c_k$ are semantically distinct, by ensuring the main action (the verb) and/or the main object (the noun) are distinct.
\Eg given the sentence ``\textit{The person closes the bin}'', the sentence ``\textit{The person opens the bin}'' is considered distinct due to the main verb being different.
However, ``\textit{The person shuts the trash can}'' is considered indistinct due to the main verb (closes/shuts) and noun (bin/trash can) both being semantically equivalent.

Next, we segment the rough clips $v_i$ and $v_j$.
To compensate for the lack of temporal bounds, we use the timestamps of surrounding narrations,  $t_{i+1}$ and $t_{i-1}$ (resp. $t_{j+1}$ and $t_{j-1}$), as upper and lower bounds for the clips.
Note that there are many possible ways of initialising these bounds (\eg in~\cite{lin2022egocentric}) which we explore further in Table \refSup{1} of the supplementary material.

To merge the clips, we randomly sample two variables $\alpha, \beta$ such that $0 \le \alpha \le 1$ is the position at which we position the first clip and $0 < \beta \le (1-\alpha)$ is its duration.
From these, we define a merging template $y$ as a rectangular function, so that $y(t) = 1$ if $\alpha < t < \alpha+\beta$, and $y(t) = 0$ otherwise. This template merges the two clips $v_i$ and $v_j$ such that:
\begin{equation}
    S_{ij}(t) = 
   \begin{cases}
       v_i((t-\alpha)/\beta) & y(t) = 1\\
       v_j(t/(1-\beta)) & y(t) = 0 \quad \& \quad t < \alpha\\
       v_j((t-\beta)/(1-\beta)) &y(t) = 0 \quad \& \quad t > \alpha
   \end{cases}
\end{equation}
where $0 \le t \le 1$ represents the time along the normalised length of any clip and $v_i(t)$ represents the clip $v_i$ at time $t$.
The template $y$ also defines the supervisory signal ($y_i$ and $y_j$) for both sentences.
For the sentence $c_i$, $y_i = y$ is a binary vector set to 1 for all locations where $S_{ij}$ contains parts of the clip $v_i$ and 0 otherwise. Analogously $y_j = \mathbbm{1}_{{|y|}}-y$, where $\mathbbm{1}_{{|y|}}$ is a vector of 1s of length $|y|$. 

Note that the above definition assumes continuous time $t$. 
Instead, we pre-extract visual video features $F$ at a fixed temporal rate from the video. 
This rate is then used to discretise the merging operation noted above. 
\subsection{Text Conditioning}
\label{subsec:text_condition}
We condition the visual features $F_{ij}$ of the merged segment $S_{ij}$ on the sentences $c_i$ and $c_j$, by matching to the corresponding sentence and contrasting from the other sentences. 
A~projection layer, $g$, is applied to extracted word features to project the sentence embeddings into the same space as the visual feature embeddings, resulting in final text features $\hat{c}_i$ with $L \times d$ dimensions where $L$ is the length of the sentence:
\begin{equation}
    \hat{c}_i = [g(f(w_{i,1})), ..., g(f(w_{i,L}))]
\end{equation}
where $w_{i,l}$ represents the $l$th word in the sentence $c_i$ and $f$ represents the text feature extractor.
To condition the visual features on $\hat{c}_i$, we 
propose a combination of both self-attention and cross-attention within successive transformer layers to learn correlations both within and between the modalities, see Figure~\ref{fig:model_full}(b).
We follow a similar approach to~\cite{lu2019vilbert} where first visual and text features undergo self-attention, before being concatenated and passed through a transformer layer using cross-attention:
\begin{equation}
    \label{eq:self_att}
    P=\mathcal{A}^s_v(F_{ij}) \quad; \quad\quad\quad  Q_i=\mathcal{A}^s_t(\hat{c}_i)
\end{equation}
where $P=(p_1, ..., p_M)$ and $Q_i=(q_{i,1}, ..., q_{i,L})$ are the outputs of the self-attention layers and $A^s$ represents a self-attention layer.
\begin{equation}
    \label{eq:cross_att}
    H_i' = \mathcal{A}^c(H_i)
\end{equation}
where $H_i = [p_1, ..., p_m, ..., p_M, q_{i,1}, ..., q_{i,l}, ..., q_{i,L}]$ (the concatenated video and textual outputs from the previous layer) and $\mathcal{A}^C$ is a cross-attention transformer layer in which keys and values are passed as inputs to the opposing modality.
The combination of self-attention and cross-attention layers, \ie equations~\ref{eq:self_att} and~\ref{eq:cross_att}, are repeated $n$ times.
We then include a residual connection from the input visual features. 
This is followed by two fully connected layers and a sigmoid function that estimates the probability of each visual input feature being the grounding for the sentence $c_i$:
\begin{equation}
    \hat{Y}_i = \mathcal{M}(F_{ij}, c_i) = \sigma \big(e(H'_i) + F_{ij}\big)
    \label{eq:estimate}
\end{equation}
where $\mathcal{M}$ defines the model \methodName, $e$ represents two fully connected layers and $\hat{Y}_i$ is the vector of estimated probabilities that the visual features ground the sentence $c_i$. 

%

\subsection{Training and Inference}
\label{sec:loss}

During training, we randomly sample clips for merging within each video in the dataset. We restrict the size of the merged clip $S_{ij}$ to a fixed size of $M$ features for batching purposes. 
We next describe the losses used to train \methodName.

The main contrasting loss, denoted collectively as $\mathcal{L}_{sep}$, is applied between the model's predicted probabilities $\hat{Y}$ and the merged clips supervisory signals $y_i$, $y_j$.
\begin{equation}
    \mathcal{L}_{sep} = (1-\beta) \cdot \text{BCE}(\hat{Y}_i, y_i) + \beta \cdot \text{BCE}(\hat{Y}_j, y_j)
    \label{sep_loss}
\end{equation}
where $\text{BCE}$ is the Binary Cross Entropy Loss, balanced by the proportion of the other sentence in the merged segment $S_{ij}$ using $\beta$ from Sec.~\ref{subsec:clip_combine} and $\hat{Y}_i$ is the vector of estimated grounding probabilities from Eq~\ref{eq:estimate}.


$\mathcal{L}_{neg}$ is applied to the negative sentence $c_k$ to prevent the method overfitting to the sampled boundaries within the merged segment $S_{ij}$.
This is then applied across video features that have been conditioned on $c_k$ where resulting probabilities should be $0$:
\begin{equation}
    \mathcal{L}_{neg} = \text{BCE}(\mathcal{M}(S_{ij},c_{k}),  0\cdot\mathbbm{1}_{|S_{ij}|})
\end{equation}
A further regularisation loss, which we term the adversarial loss $\mathcal{L}_{adv}$, is used on the non-conditioned visual features (i.e. without any text conditioning). The adversarial loss enforces that \methodName{} uses both the textual query and the visual frames to retrieve moments.
This follows related works which find that temporal sentence grounding models can perform well with no visual inputs~\cite{otani2020uncovering,yuan2021closer}.
Using this loss, visual features with no text are conditioned to produce a vector with maximal uncertainty (i.e. $0.5\cdot\mathbbm{1}_{|S_{ij}|})$).
\begin{equation}
    \mathcal{L}_{adv} = \text{BCE}(\mathcal{M}(S_{ij}),  0.5\cdot\mathbbm{1}_{|S_{ij}|})
\end{equation}
The final loss to train the model is given as:
\begin{equation}
    \mathcal{L}=\mathcal{L}_{sep} + \lambda_{neg} \cdot \mathcal{L}_{neg} + \lambda_{adv} \cdot \mathcal{L}_{adv} 
    \label{loss_sum}
\end{equation}
where $\lambda_{\{neg, adv\}}$ are weights used to balance the different losses.


At inference, the model considers a single sentence and is applied to the full untrimmed length of a single video, no longer restricting the number of features. 
The model outputs the probability of grounding the sentence across the full extent of the video, not merged segments.
To predict the grounding, we normalise the predictions between $0$ and $1$ and consider a threshold $\epsilon$ to convert predictions into a grounding such that ${\forall \hat{t}: \hat{t}^s_i \le \hat{t} \le \hat{t}^e_i \iff \hat{Y}_i(\hat{t}) > \epsilon}$.
Predicted groundings ($\hat{t}^s_i$, $\hat{t}^e_i$) are ranked by each one's maximum prediction value and returned as a ranked list.
We then calculate the temporal IoU of the highest ranked grounding against the ground truth ($t^s_i$, $t^e_i$) to report performance.


\section{Experiments}
\label{sec:experiments}

\myparagraph{Metrics} We use the standard metric for \taskNameMini, Recall@K (R@K) with Intersection Over Union~(IoU) at threshold $\theta$ (IoU=$\theta$).
As in~\cite{zhang2021natural}, we evaluate ${\theta\in\{0.1,0.3,0.5\}}$ and K=1. 
We also report Mean Recall@1 (mR@1) metric introduced in~\cite{barrios2023multimodal} as the average across $\theta$.

\myparagraph{Baselines} We select VSLNet~\cite{zhang2021natural} and its implementation in~\cite{grauman2022ego4d} as our baseline. For fair comparison, we use the same Omnivore visual features and BERT text features for training this baseline as we use for \methodName{}.
This is one of two baselines used for the NLQ task in Ego4D and has shown high performance on grounding tasks in previous datasets.
We train VSLNet using artificially generated start and end times based on the rough timestamps.
We also include a random baseline which samples a random segment of any length from the video to demonstrate the challenge of the test sets.


\myparagraph{Datasets} We use two datasets: Ego4D~\cite{grauman2022ego4d} and EPIC-Kitchens-100~\cite{damen2022rescaling}. Both datasets have published a mapping from open vocabulary to closed vocabulary, using manual and automatic clustering. For each dataset, we consider all verbs/nouns within the same ‘closed
vocabulary’ cluster to be semantically equivalent (i.e. pan
and saucepan are both contained within the pan cluster so
are considered semantically equivalent). These are used to measure the semantic equivalence of sentences.

\myparagraph{Ego4D}: We consider a subset of videos from~\cite{grauman2022ego4d} covering various scenarios: \{\textit{cooking, carpenter, scooter mechanic, car mechanic, gardener, farmer}\}.
Annotations from the hand-object interaction challenge are used as ground truth in the val/test splits, as these videos contain human-labelled start and end times with corresponding sentences.
Due to the nature of the Ego4D videos, there are often a number of repeated or semantically similar sentences within the videos.
Currently, \taskNameMini{} approaches assume there is only one positive grounding moment for each sentence.
To align with the grounding task, any repeated or semantically indistinct sentences were excluded from the val/test sets.
This resulted in a dataset with 462/96/223 videos and 197k/1.9k/4.4k sentences in train/val/test. When using the val/test sets, only full videos are used, no clip merging takes place.

\myparagraph{EPIC-Kitchens-100}: 
We use all 495 videos in the EPIC-Kitchens-100 train set to make up our train/val/test splits.
We use the narration annotations for training and the manually labelled start-end times for evaluation. We similarly remove semantically equivalent sentences, and the resulting dataset has 
332/50/110 videos and 45k/1.8k/3.8k sentences in train/val/test respectively. 


\myparagraph{Features}
BERT~\cite{devlin2019bert} is used to generate features for the sentences, trained on BookCorpus~\cite{zhubookcorpus2015} and English Wikipedia, generating 768-length features for each word.
Visual features are obtained from the video head of the Omnivore~\cite{girdhar2022omni} model (SwinL for Ego4D, SwinB for EPIC-Kitchens) trained on Kinetics-400 and ImageNet-1K jointly and are of length 1536 for Ego4D and 1024 for EPIC-Kitchens.
Each visual feature represents 32 frames of the video, and these are generated with a stride of 16 frames.
As Ego4D videos have an FPS of 30, the feature rate is $\sim$1.87 per second.
EPIC-Kitchens videos are downsampled to 30 FPS to match this.

\begin{table}[t]
\centering
\begin{minipage}{0.49\textwidth}
\centering
\resizebox{\columnwidth}{!}{%
\begin{tabular}{lrrrr}
\hline
&  IoU=0.1                        & IoU=0.3             & IoU=0.5   & mR\\ \hline
Random Baseline               & 0.47                            & 0.09                  & 0.02   & 0.19\\
VSLNet~\cite{zhang2021natural}               & 7.32                            & 3.16                  & 1.35  & 3.94 \\
Cli$\mathcal{M}$er                     & \textbf{9.68}                            & \textbf{5.03}                  & \textbf{2.24}  & \textbf{5.65}\\  \hline
\end{tabular}%
}
\caption{Results for R@1 against baselines on Ego4D~\cite{grauman2022ego4d}}
\label{tab:main_ego4d}
\end{minipage}
\begin{minipage}{0.49\textwidth}
\centering
\resizebox{\columnwidth}{!}{%
\begin{tabular}{lrrrr}
\hline           
&  IoU=0.1                        & IoU=0.3             & IoU=0.5 & mR \\ \hline
Random Baseline& 0.78     & 0.13    & 0.03   &  0.31 \\
VSLNet~\cite{zhang2021natural}         & 19.32    &  8.76    & 3.90   &  10.66\\
\methodName         & \textbf{22.20}    & \textbf{11.57}    & \textbf{5.25}   & \textbf{13.01}\\ \hline
\end{tabular}%
}
\caption{Results for R@1 against baselines on EPIC-Kitchens~\cite{damen2022rescaling}}
\label{tab:main_epic}
\end{minipage}
\end{table}

\begin{figure*}[t]
    \centering
    \includegraphics[width = \textwidth]{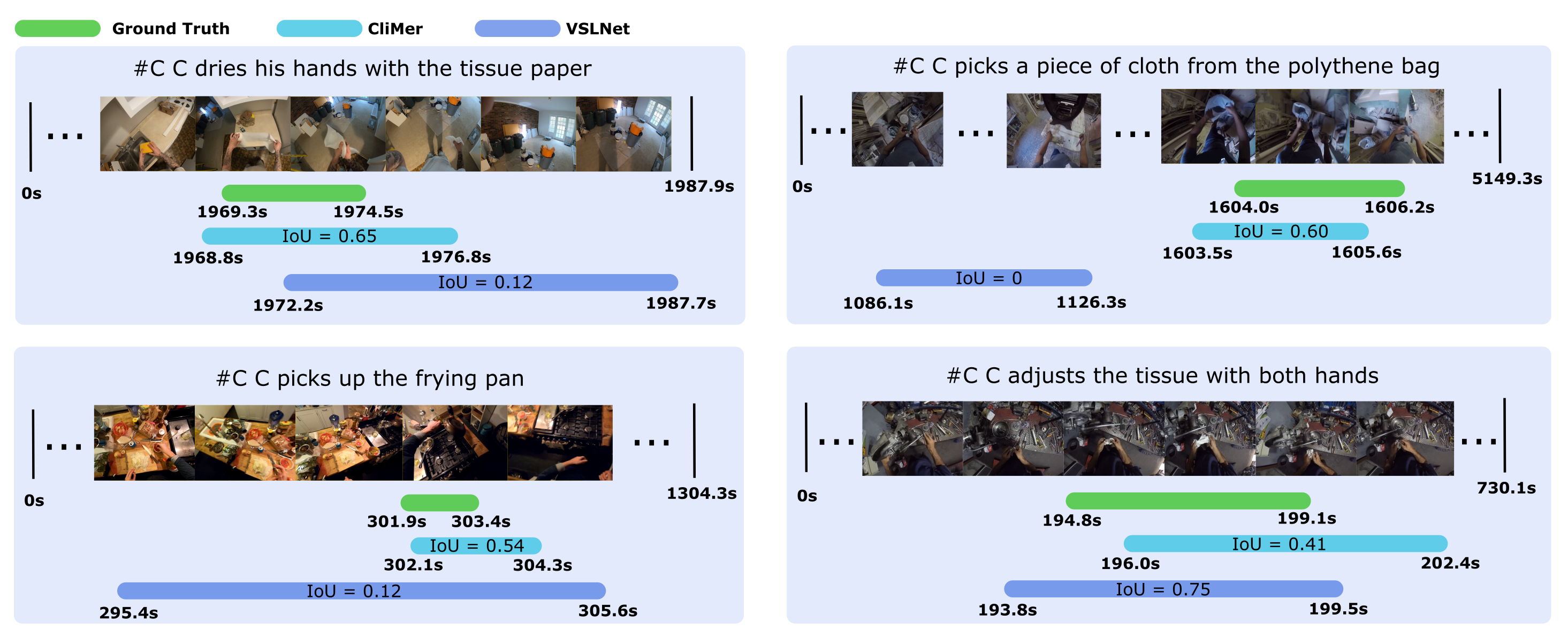}
    \caption{Examples from Ego4D comparing \methodName{} and VSLNet with ground truth.}
    \label{fig:qual_results}
        \vspace*{-12pt}
\end{figure*}

\myparagraph{Model}
The textual features and visual features are first projected to the common size of $2048$ dimensions.
For Ego4D, the number of blocks of self and cross-attention before the final self-attention block is set to $n=2$.
For EPIC-Kitchens, we switch the text attenuation module with a simpler one. 
In this case, the visual features undergo two self-attention layers before a Hadamard product is used to attenuate them with mean-pooled text features instead of a cross attention layer.
We ablate this design choice in Table~\ref{tab:text_att} and show it is more suited to the simpler sentences within EPIC-Kitchens.
%

All attention layers contain 6 heads.  
The number of visual features during training is set to $M=20$. 
We use the Adam optimiser with a learning rate of $10^{-5}$ and batch size of 32. We use a dropout value of 0.3 and a value of 0.5 for the text feature projection. 
The model is trained for 100 epochs and the epoch with the best validation result is selected to report performance on the test set (\ie we do not optimise on the test set).
We set the weights $\lambda_{neg}=\lambda_{adv}=1$ and ablate these losses.
The prediction threshold, $\epsilon$ is empirically chosen as 0.8 for Ego4D and 0.6 for EPIC-Kitchens (see Figure~\ref{fig:pred_threshold}).

\subsection{Results}
\label{subsec:results}

We first show results of \methodName{} on Ego4D in Table~\ref{tab:main_ego4d}.
The random baseline shows the challenge of this dataset.
Although VSLNet has been adapted for long videos as described in~\cite{grauman2022ego4d}, \methodName{} outperforms it across all metrics.

Table~\ref{tab:main_epic} includes the results on EPIC-Kitchens.
Random performance continues to be low for EPIC-Kitchens, and both methods perform considerably better than it.
Once again, \methodName{} achieves higher performance across all metrics. We believe the contrastive learning enabled by our clip merging method allows the model to learn to locate corresponding video segments more effectively when given an input sentence.

Figure \ref{fig:qual_results} shows qualitative examples comparing \methodName{} with VSLNet~\cite{zhang2021natural}.
VSLNet often over-predicts start/end times leading to poor IoU scores when compared to the ground truth. An extreme example of this can be seen in the top right predicting a $\sim$40s clip for a $\sim$2s ground truth segment.
We find that \methodName{} is able to match the ground truth more closely, despite the fine-grained nature of the sentences.
In the bottom right example, \methodName{} continues to predict the part of the video where the tissue is visible but is no longer being adjusted.


\begin{table}[t]
\centering
\resizebox{0.6\columnwidth}{!}{%
\begin{tabular}{ccc|cccc}
\hline
 $c_j$ & $\mathcal{L}_{neg}$ & $\mathcal{L}_{adv}$  & IoU=0.1                        & IoU=0.3             & IoU=0.5  & mR\\ \hline
                  $\checkmark$ & $\checkmark$ & -             & 9.57                       & 4.46      & 2.13 & 5.39\\ 
                  $\checkmark$ & - & $\checkmark$             & 8.61                       & 3.62      & 1.58 & 4.60 \\
                  -  & $\checkmark$ & $\checkmark$             & 8.88                            & 4.35                  & 1.90  & 5.04\\ \hline
                  $\checkmark$ & $\checkmark$ & $\checkmark$             & \textbf{9.68}                 & \textbf{5.03}       & \textbf{2.24} & \textbf{5.65} \\ \hline
\end{tabular}}
\caption{Impact of removing each loss during training on \methodName{}.}
\label{tab:ablation_losses}
\end{table}
\begin{table}[t]
\centering
\resizebox{0.6\columnwidth}{!}{%
\begin{tabular}{cc|cccc}
\hline
 $\lambda_{neg}$ & $\lambda_{adv}$ & IoU=0.1                        & IoU=0.3             & IoU=0.5  & mR\\ \hline
                  0.5 & 1             & 8.93                       & 4.12      & 1.95 & 5.00\\ 
                  1 & 0.5             & 8.88                       & 4.49      & 2.08 & 5.15 \\
                  1 & 0             & 9.57                         & 4.46      & 2.13 & 5.39\\
                  0 & 1               & 8.61 & 3.62 & 1.58 & 4.60\\
                  1 & 1             & \textbf{9.68}                 & \textbf{5.03}       & \textbf{2.24} & \textbf{5.65} \\\hline
\end{tabular}}
\caption{Effect of changing the loss weights on \methodName{}.}
\label{tab:ablation_loss_weights}
\end{table}

\begin{table}[h!]

\begin{minipage}{0.27\textwidth}
    \centering
    \includegraphics[width=\textwidth]{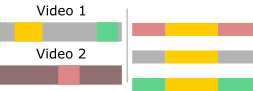}
\end{minipage} \hfill
\begin{minipage}{0.72\textwidth}
    \resizebox{\columnwidth}{!}{%
\begin{tabular}{cccccc}
\hline
 Merged Segment & Hard Negatives  & IoU=0.1                        & IoU=0.3             & IoU=0.5  & mR\\ \hline
$\checkmark$ & - & 7.28                            & 3.09                  & 1.08 & 3.82\\ 
- & $\checkmark$ & 7.23                            & 4.42                  & 2.20  &  4.62\\ 
$\checkmark$ & $\checkmark$ & \textbf{9.68}                            & \textbf{5.03}                  & \textbf{2.24}  & \textbf{5.65}\\ \hline
\end{tabular}%
}
\end{minipage}
\caption{Performance on Ego4D with ablations on merging clips to produce the supervision signal and sampling clips from the same video as hard negatives. (Left inset) We show how each row corresponds to the merged segment creation.}
\vspace*{-12pt}
\label{tab:merge_ablation}
\end{table}

\subsection{Ablation Studies}
\label{subsec:ablation}


\myparagraph{Losses}
In Table~\ref{tab:ablation_losses} we showcase that each loss contributes to the model's performance, with performance decreasing when any are omitted.
Specifically,  
removing the loss of the second sentence~($c_j$) from $\mathcal{L}_{sep}$ removes the contrastive learning aspect of the model, leading to performance decreases, albeit still being able to achieve reasonable performance via grounding the single sentence only.
Removing $\mathcal{L}_{adv}$ has the least impact, being a regularisation loss. 
Removing $\mathcal{L}_{neg}$ leads to the largest performance drop, as the model otherwise assumes that all sentences can be grounded resulting in an increase of false positives. 

We also adjust the weights of the negative and adversarial losses ($\lambda_{neg}$ and $\lambda_{adv}$ from Eq.~\ref{loss_sum}) in the loss function to examine the effect this has on model performance for Ego4D.
The results in Table~\ref{tab:ablation_loss_weights} show that decreasing either weights results in reduced performance across all metrics. 
This demonstrates the importance of each loss having equal weighting to the main separation loss, providing complementary signals during training.


\myparagraph{Clip Merging} We investigate the impact of merging segments.
We replace merging with using a single clip with surrounding background. 
The background is selected to make up between 25\% and 75\% of the video at random, retaining the value of $M=20$ for the number of features.
The losses are retained as before however the weighting of loss $\mathcal{L}_{sep}$ is doubled to compensate for the lack of contribution from $c_j$.

From results in Table \ref{tab:merge_ablation} (row 2) without merging, performance drops.
The decrease at IoU=0.1 in particular shows that the model without merging struggles to learn the general alignment between sentences and video, likely due to the lack of contrastive learning and the decreased visual diversity found in continuous segments compared to merged segments. The model trained without merging frequently misses the ground truth moment entirely.
The results show that clip merging increases robustness to changes in clip positions and lengths in full videos. 




\myparagraph{Hard Negative Sampling}
Table~\ref{tab:merge_ablation} also compares clips sampled as hard negatives from the same video against when they are sampled from random videos.
A significant performance drop is shown in the latter case.
Hard negatives force the model to distinguish between the actions occurring within the videos as the visual environments of the two clips are similar.


\begin{table}[t]
\begin{minipage}{0.75\textwidth}
    \centering
    \resizebox{\columnwidth}{!}{%
    \begin{tabular}{crrrrrrrrr}
    \hline
         & \multicolumn{4}{c}{Ego4D}& &\multicolumn{4}{c}{EPIC-Kitchens} \\ \cline{2-5} \cline{7-10}
         &  IoU=0.1 & IoU=0.3 & IoU=0.5 & mR && IoU=0.1 & IoU=0.3 & IoU=0.5 & mR \\ \hline
    Hadamard Prod. Cross Att. & 9.29 & 4.30 & 2.04 & 5.21 && \textbf{22.20} & \textbf{11.57} & 5.25 & \textbf{13.01} \\
    Learned Cross Att.    & \textbf{9.68} & \textbf{5.03} & \textbf{2.24} & \textbf{5.65} && 19.79 & 10.58 & \textbf{5.46} & 11.94 \\
    \hline
    \end{tabular}%
    }
    \end{minipage}\hfill
\begin{minipage}{0.24\textwidth}
    \centering
    \includegraphics[width=\textwidth]{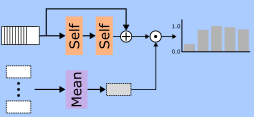}
\end{minipage} 
    \caption{Ablation comparing the type of text attenuation used: Cross-attention as in Figure~\ref{fig:model_full}(b) or using a Hadamard Product (shown right).}
    \label{tab:text_att}
\end{table}
\begin{figure}[t]
    \begin{minipage}{0.48\textwidth}
        \centering
        \includegraphics[width=\textwidth]{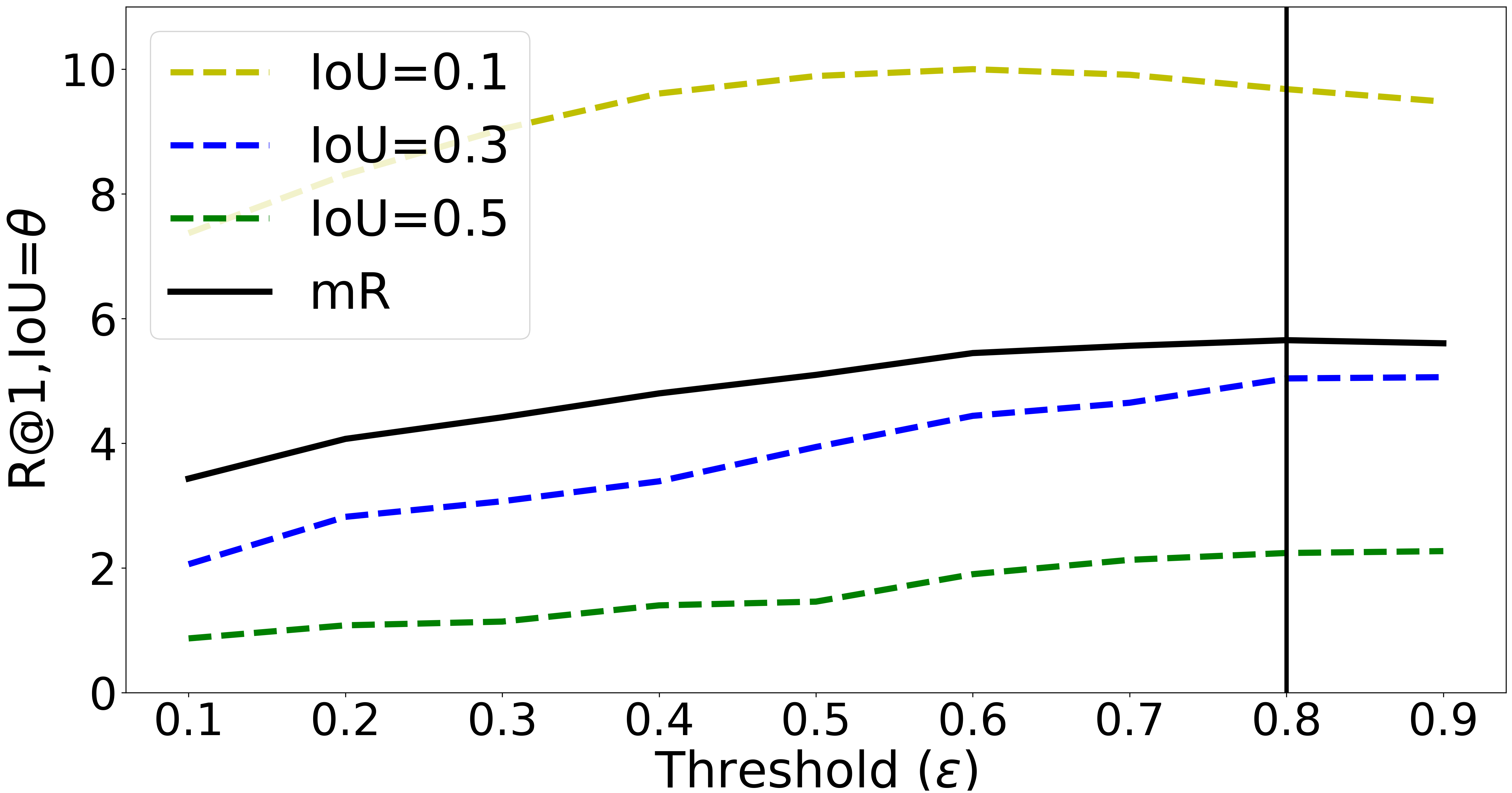}
    \end{minipage} \hfill
    \begin{minipage}{0.48\textwidth}
        \centering
        \includegraphics[width=\textwidth]{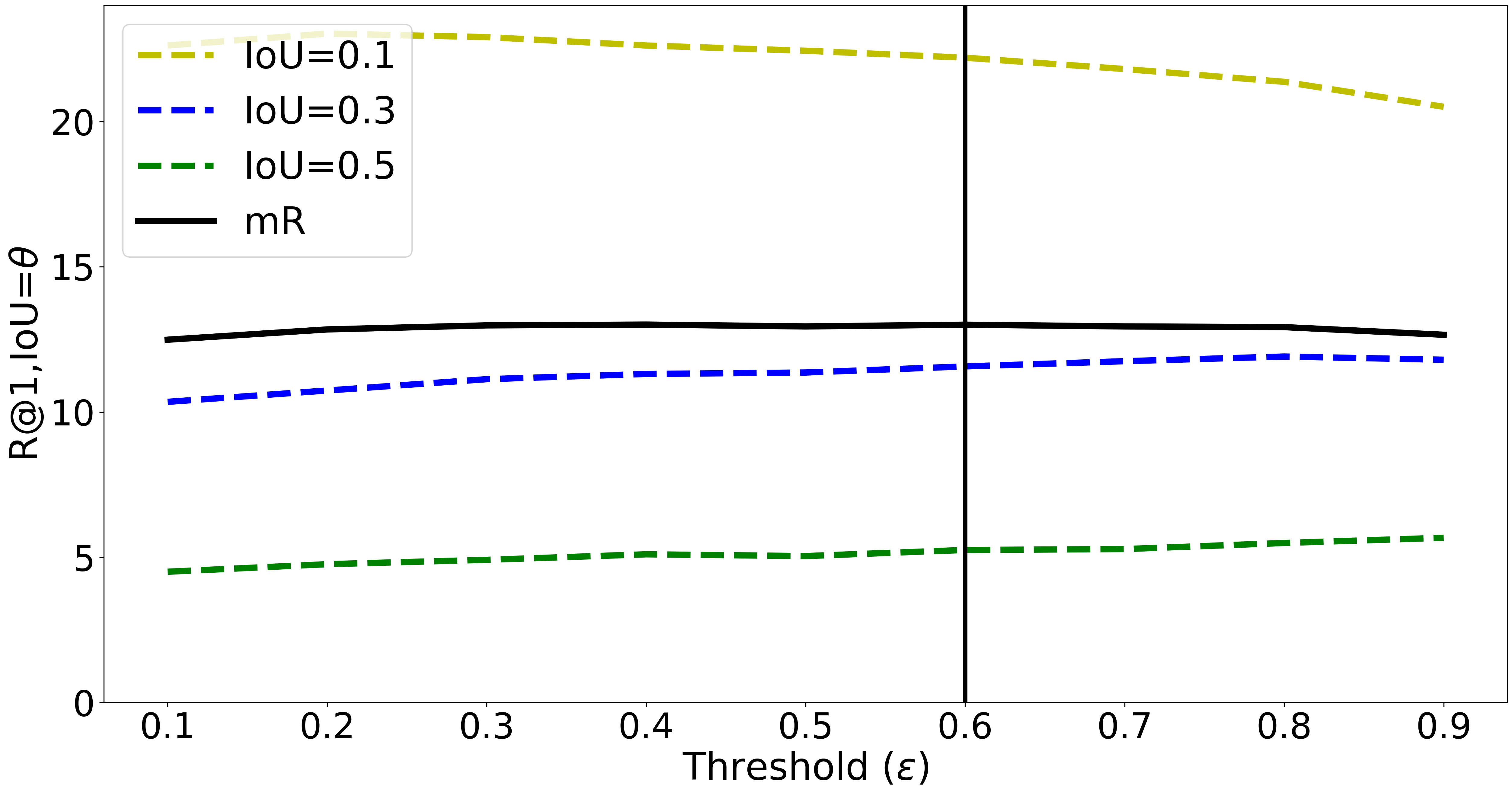}
    \end{minipage} \hfill
    \caption{\methodName{} performance as the prediction threshold, $\epsilon$, is varied on Ego4D (Left) and EPIC-Kitchens (Right). The vertical lines indicate our thresholds of 0.8 and 0.6.}
    \label{fig:pred_threshold}
\end{figure}

\myparagraph{Text Attenuation}
Table~\ref{tab:text_att} evaluates the choice of text conditioning module, comparing using Cross Attention (see Section~\ref{subsec:text_condition}) with the Hadamard product.
The simpler module which we use for EPIC-Kitchens works best due to the smaller dataset and shorter sentences (avg. $3.0$ words) vs. Ego4D (avg. $8.4$ words).

\myparagraph{Prediction Threshold}
Figure \ref{fig:pred_threshold} displays the effect of varying the value of the prediction threshold $\epsilon$.
The threshold $\epsilon$ has more impact on Ego4D, however the figure
demonstrates that the model is reasonably robust to the choice of $\epsilon > 0.5$. 

\section{Conclusion}
\label{sec:conclusion}

In this work we have explored \taskName{} for long-form egocentric datasets Ego4D and EPIC-Kitchens.
To overcome the need for full supervision, we propose to merge clips from the same video from rough narration timestamps.
Our method, \methodName, trains using contrastive learning with text-conditioning over sentences.
Results demonstrate the ability of \methodName{} to tackle long videos at test time, grounding fine-grained sentences in egocentric videos which can be over an hour in length. 

\paragraph{Acknowledgments.} K Flanagan is supported by UKRI (Grant ref EP/S022937/1) CDT in Interactive AI \& Qinetiq Ltd via studentship CON11954. D Damen is supported by EPSRC Fellowship UMPIRE (EP/T004991/1) \& EPSRC Program Grant Visual AI (EP/T028572/1). 

\bibliography{egbib}

\end{document}